\documentclass[final,5p,times,twocolumn,authoryear]{elsarticle}

\usepackage{amssymb}
\usepackage{amsmath}
\usepackage{tikz}
\usepackage{hyperref}
\usepackage{url}
\usetikzlibrary{positioning,arrows.meta,fit,calc}

\journal{Medical Image Analysis}

\begin{document}

\begin{frontmatter}



\title{IMA-MoE: An Interpretable Modality-Aware Mixture-of-Experts Framework for Characterizing the Neurobiological Signatures of Binge Eating Disorder} 

\fntext[fn1]{Contribute equally to this work.}
\cortext[cor1]{Corresponding authors.}

\author[label1]{Lin Zhao\fnref{fn1}\corref{cor1}}
\ead{lin.zhao.1@njit.edu}
\author[label2]{Qiaohui Gao\fnref{fn1}}
\author[label1]{Elizabeth Martin}
\author[label3]{Kurt P. Schulz}
\author[label3]{Tom Hildebrandt}
\author[label3]{Robyn Sysko}
\author[label4]{Tianming Liu}
\author[label1]{Xiaobo Li\corref{cor1}}
\ead{xiaobo.li@njit.edu}
\affiliation[label1]{
organization={Department of Biomedical Engineering, New Jersey Institute of Technology},
city={Newark},
postcode={07102}, 
state={NJ},
country={USA}
}

\affiliation[label2]{
organization={College of Engineering, Northeastern University},
city={Boston},
postcode={02115}, 
state={MA},
country={USA} 
}

\affiliation[label3]{
organization={Department of Psychiatry, Icahn School of Medicine at Mount Sinai},
city={New York},
postcode={30602}, 
state={NY},
country={10029}
}

\affiliation[label4]{
organization={School of Computing, University of Georgia},
city={Athens},
postcode={30602}, 
state={GA},
country={USA}
}

\begin{abstract}

Binge eating disorder (BED) is the most prevalent eating disorder. However, current diagnostic frameworks remain largely grounded in symptom-based criteria rather than underlying biological mechanisms, thereby limiting early detection and the development of biologically-informed interventions. Emerging studies have begun to investigate the neurobiological signatures of BED, yet their findings are often difficult to generalize due to the reliance on hypothesis-driven parametric models, single-modality analyses, and limited data diversity. Therefore, there is a critical need for advanced data-driven frameworks capable of modeling multimodal data to uncover generalizable and biologically meaningful signatures of BED. In this study, we propose the Interpretable Modality-Aware Mixture-of-Experts (IMA-MoE), a novel architecture designed to integrate heterogeneous neuroimaging, behavioral, hormonal, and demographic measures within a unified predictive framework. By encoding each measure as a distinct token, IMA-MoE enables flexible modeling of cross-modal dependencies while preserving modality-specific characteristics. We further introduce a token-importance mechanism to enhance interpretability by quantifying the contribution of each measure to model predictions. Evaluated on the large-scale Adolescent Brain Cognitive Development (ABCD) dataset, IMA-MoE demonstrates superior performance in differentiating BED from healthy controls compared with baseline methods, while revealing sex-specific predictive patterns, with hormonal measures contributing more prominently to prediction in females. Collectively, these findings highlight the promise of interpretable, data-driven multimodal modeling in advancing biologically-informed characterization of BED and facilitating more precise and personalized interventions in neuropsychiatric disorders.

\end{abstract}




\end{frontmatter}



\section{Introduction}

Binge eating disorder (BED) is the most prevalent eating disorder characterized by recurrent episodes of consuming large quantities of food accompanied by a sense of loss of control~\citep{dingemans2002binge,qian2013prevalence}. Individuals with BED often experience marked psychological distress and face elevated risks for obesity, metabolic dysfunction, and reduced quality of life~\citep{giel2022binge}. Despite its high prevalence and substantial health burden, the diagnosis of BED remains largely grounded in symptom-based taxonomies that emphasize observable behaviors and self-reported experiences; although these criteria are clinically useful for classifying established disorders, their limited connection to underlying biological mechanisms constrains early detection and the development of risk-informed, personalized intervention strategies.

Over the past several decades, growing evidence has begun to elucidate the biological mechanisms underlying BED, suggesting that structural brain changes, functional brain network alterations, and endocrine factors are each associated with its clinical manifestation~\citep{murray2014hormonal,kober2018potential,pasquale2024reward}. Structural MRI (sMRI) studies have identified morphological alterations in BED, including increased gray matter volume in reward-related brain regions such as medial orbitofrontal cortex and left nucleus accumbens~\citep{schafer2010regional,abdo2020relationship}. Diffusion Tensor Imaging (DTI) studies reveal microstructural dysregulation such as increased axial diffusivity and fractional anisotropy in fronto-limbic and temporoparietal pathways~\citep{estella2020brain}, and increased connectivity in minor forceps, corpus callosum, cingulate gyrus and superior longitudinal fasciculus~\citep{hartogsveld2022volume}. Functionally, task-based functional MRI (tfMRI) studies have revealed elevated responsivity of reward network to food cues presentation~\citep{weygandt2012diagnosing, lee2017impaired} and altered brain network topology~\citep{martin2025altered}. Resting-state functional connectivity analyses support network-level dysfunction, such as decreased functional connectivity in the striatum\citep{haynos2021resting} and abnormalities in functional synergy between reward and executive control networks~\citep{griffiths2024functional}. In parallel, these neural mechanisms are inextricably linked to the endocrine system, where pubertal surges in hormones appear to activate genetic risks and alternations in widespread brain networks, thus exacerbating binge-eating vulnerability in females~\citep{klump2017significant,culbert2021narrative,martin2025distinct}.

Despite these advances, current research on BED remains fragmented, and mechanistic conclusions are often difficult to generalize across studies. Findings across structural, functional, and hormonal domains are frequently heterogeneous, with some studies reporting opposing or null effects even when probing similar brain regions~\citep{schafer2010regional,abdo2020relationship,hagan2021subcortical}. One important contributor to this discrepancy is the predominance of parametric, hypothesis-driven models that rely on strong assumptions about variable independence and linear effects; while such approaches allow controlled statistical inference, they may oversimplify the interdependent biological processes across neural, behavioral, and endocrine systems that jointly shape disorder expression. In addition, many studies are constrained by modest sample sizes, single data modality, cross-sectional designs, heterogeneous data-inclusion criteria, and varying degrees of comorbidity control (e.g., obesity, mood disorders, medication use), all of which can bias effect estimates and impede generalization. Consequently, there is a critical need for advanced data-driven analytical frameworks capable of modeling multimodal, interdependent biological data using large-scale and demographically diverse datasets to generate more stable and generalizable insights into BED-related neurobiology.

Recently, the rapid development of artificial intelligence (AI) methods has offered a promising avenue to address these challenges by enabling integrative modeling of complex, multimodal biological data in a data-driven manner. Multimodal transformers and Mixture-of-Experts (MoE) frameworks have emerged to synthesize heterogeneous data sources ranging from high-dimensional neuroimaging (sMRI, fMRI, DTI) to genomic profiles and clinical assessments into unified predictive models~\citep{yun2024flex,zhuang2025multimodal,shaik2025multi,lyu2025oblique}. For instance, Multi-modal Imaging Genomics Transformer (MIGTrans) utilizes cross-attention mechanisms to integrate genomics with structural and functional imaging data to identify Schizophrenia~\citep{shaik2025multi}. Similarly, in neurodegenerative research, MoE-based architectures such as Oblique Genomics Mixture of Experts(OG-MoE)~\citep{lyu2025oblique} dynamically route information to specialized "experts," to improve the diagnosis of Alzheimer’s disease progression based on multimodal dataset~\citep{yun2024flex,burns2025flexible,ding2025denseformer}. However, challenges remain in applying those data-driven frameworks to BED. Integrating highly heterogeneous data sources is nontrivial, as high-dimensional neuroimaging measures and lower-dimensional hormonal or demographic measures differ substantially in structure and characteristics, making it challenging to achieve a balanced and integrated multimodal representation while preserving modality-specific information. In addition, predictive performance alone does not readily yield biological insight. Limited interpretability in those frameworks makes it difficult to determine which measures drive model-derived prediction, constraining their value for mechanistic understanding.

To address these challenges, we propose the Interpretable Modality-Aware Mixture-of-Experts (IMA-MoE), a novel architecture designed to integrate heterogeneous neuroimaging, behavioral, hormonal, and demographic measures while characterizing the neurobiological signatures of BED. Since these measures differ substantially in dimensionality, structure, and characteristics, we employ measure-specific encoders separately designed for high-dimensional vector inputs and lower-dimensional scalar variables to generate structured embeddings. Each measure is then represented as a distinct token, enabling the model to preserve measure-specific information. These token-based embeddings are refined through a transformer encoder that enables cross-modal interactions, followed by a Mixture-of-Experts (MoE) module that dynamically routes tokens through specialized processing pathways. Importantly, we incorporate a lightweight token-importance module that quantifies the contribution of each measure to the model’s output, thereby enhancing interpretability. We evaluated the IMA-MoE framework on the large-scale Adolescent Brain Cognitive Development (ABCD) Study dataset~\citep{casey2018adolescent}. Experimental results demonstrate that IMA-MoE effectively handles the heterogeneity in different measures and achieves superior performance in differentiating BED from healthy controls compared with baseline methods. Notably, the model reveals sex-specific differences in both predictive performance and feature attribution, with hormonal measures contributing more strongly to predictions in females, who also show higher classification performance (91.67\%) than males (72.73\%), consistent with prior literature. Overall, this work highlights the broader promise of data-driven, interpretable multimodal modeling for advancing a more comprehensive understanding of the underlying neurobiological signatures of BED.

\section{Method}

\begin{figure*}[t]
\begin{center}
\includegraphics[width=1\linewidth]{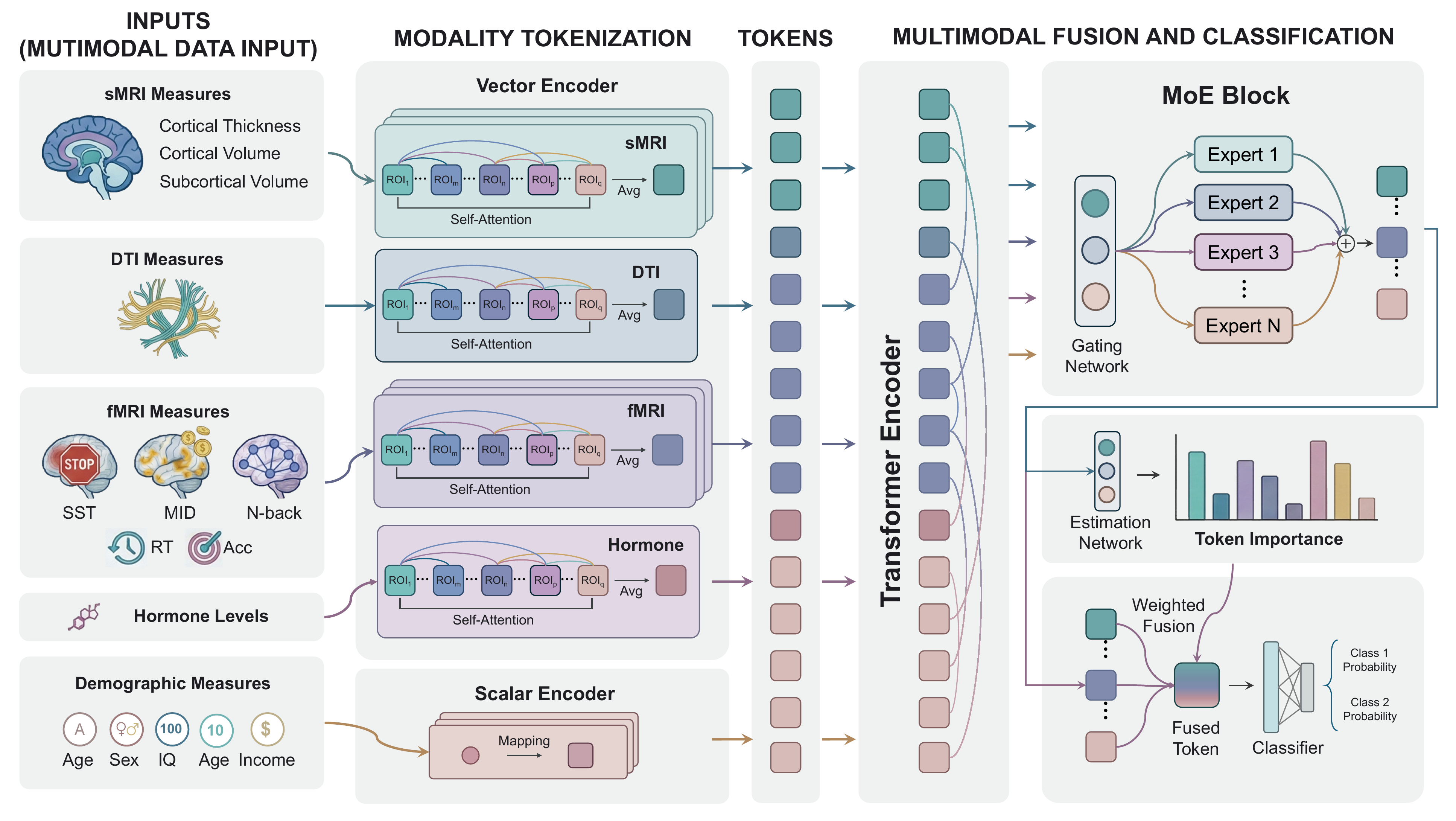}
\end{center}
\caption{Overview of the proposed Interpretable Modality-Aware Mixture-of-Experts (IMA-MoE) framework. Multimodal inputs including sMRI, DTI, fMRI, hormonal, and demographic measures are first processed through modality-specific encoders. High-dimensional measures are encoded using vector encoders with self-attention mechanisms, while lower-dimensional scalar variables are mapped through a scalar encoder. Each measure is then represented as a distinct token and passed to a transformer encoder to model cross-modal dependencies. The resulting token representations are dynamically routed through a MoE block via a gating network for specialized processing. A token-importance module quantifies the contribution of each measure to the final prediction, enabling interpretable weighted fusion. The fused representation is subsequently fed into a classifier to generate probabilistic outputs for BED classification.}
\label{fig:fig1}
\end{figure*}

IMA-MoE is designed to integrate heterogeneous sMRI, DTI, fMRI, behavioral, hormonal and demographic measures that differ substantially in dimensionality, structure, and characteristics. To preserve modality-specific information while enabling rich cross-modal interactions, the proposed architecture (Figure~\ref{fig:fig1}) represents each measure as a separate token (Section~\ref{sec:token}). These tokens are subsequently refined through a transformer encoder (Section~\ref{sec:trans}), followed by a MoE module (Section~\ref{sec:moe}) and a lightweight token-importance module (Section~\ref{sec:tim}). An overview of the proposed framework is illustrated in Figure~\ref{fig:fig1}.

\subsection{Problem Formulation}

Each participant is characterized by multimodal measurements comprising 
$M$ vector-based measures
$\mathbf{x}^{(1)},\dots,\mathbf{x}^{(M)} \in \mathbb{R}^{d_m}$
(e.g., sfMRI measures)
and $K$ scalar-valued measures 
$s^{(1)},\dots,s^{(K)} \in \mathbb{R}$
(e.g., demographic measures).
The multimodal input can therefore be expressed as
\begin{equation}
\mathcal{X} =
\big\{
\mathbf{x}^{(1)},\dots,\mathbf{x}^{(M)},\;
s^{(1)},\dots,s^{(K)}
\big\}.
\end{equation}

We formulate the data-driven characterization of the neurobiological signatures associated with BED as a supervised classification problem. Specifically, a neural network model $f_\theta(\cdot)$ is trained to predict the diagnostic label $y \in \{0,1\}$ from heterogeneous multimodal inputs $\mathcal{X}$,
\begin{equation}
\hat{y} = f_\theta(\mathcal{X}),
\end{equation}
where $\hat{y}$ denotes the predicted probability of BED.

Beyond classification, we seek to quantify the relative contribution of each measure to the model's decision-making process. To this end, we jointly learn a measure-level importance mapping
\begin{equation}
\boldsymbol{\alpha} = g_\theta(\mathcal{X}) \in \mathbb{R}^{M+K},
\end{equation}
where each element $\alpha_i$ represents the contribution of the corresponding measure to the final prediction. 

\subsection{Measure Tokenization}
\label{sec:token}
Given the multimodal input $\mathcal{X}$, a straightforward approach is to concatenate all measures into a single feature vector and apply a standard classifier for prediction. However, such a representation disregards modality-specific structure and treats heterogeneous measures as homogeneous inputs. In particular, high-dimensional neuroimaging measures may dominate lower-dimensional demographic measures, potentially obscuring modality-specific contributions that are biologically meaningful. To address this limitation, we adopt a tokenization strategy, in which each measure is mapped to an independent token. This representation preserves measure-level information while enabling subsequent transformer-based attention mechanisms to model cross-modal interactions in a structured manner, without allowing high-dimensional features to eclipse clinically relevant scalar measures.

\subsubsection{Vector Encoder}

Vector-based measures may consist of structured observations defined over spatial regions (e.g., cortical thickness across cortical regions) or compact biological profiles (e.g., hormonal measures such as estradiol, DHEA, and estrone). Our objective is to encode each vector-based measure into a unified measure token while explicitly modeling the dependencies among its constituent elements (e.g., spatial regions) prior to multimodal integration.

Let $\mathbf{x}^{(m)} \in \mathbb{R}^{L_m}$ denote the $m$-th vector measure, where $L_m$ 
represents the number of elements associated with that measure. To enable interactions across modalities with differing dimensionalities, 
each element of $\mathbf{x}^{(m)}$ is projected into a latent space 
of dimension $d$:
\begin{equation}
\mathbf{h}^{(m)}_i = \mathbf{W}^{(m)}_v x^{(m)}_i + \mathbf{b}_v,
\qquad
\mathbf{h}^{(m)}_i \in \mathbb{R}^{d}.
\end{equation}

This projection transforms the vector-based measure into a latent sequence
$\mathbf{H}^{(m)} = (\mathbf{h}^{(m)}_1,\dots,\mathbf{h}^{(m)}_{L_m}) 
\in \mathbb{R}^{L_m \times d}$.
Positional embeddings are incorporated to preserve element-wise correspondence:
\begin{equation}
\tilde{\mathbf{h}}^{(m)}_i = \mathbf{h}^{(m)}_i + \mathbf{p}_i.
\end{equation}

A lightweight transformer encoder is then applied to model intra-modality 
interactions among the elements:
\begin{equation}
\mathbf{Z}^{(m)} = \mathrm{TransEnc}\!\left(
\tilde{\mathbf{h}}^{(m)}_1,\dots,\tilde{\mathbf{h}}^{(m)}_{L_m}
\right),
\qquad
\mathbf{Z}^{(m)} \in \mathbb{R}^{L_m \times d}.
\end{equation}

This step enables the encoder to capture dependencies between elements within 
each measure. 
Finally, the encoded sequence is aggregated via mean pooling to yield a 
measure-level token:
\begin{equation}
\mathbf{t}^{(m)} =
\frac{1}{L_m} \sum_{i=1}^{L_m} \mathbf{Z}^{(m)}_i,
\qquad
\mathbf{t}^{(m)} \in \mathbb{R}^{d}.
\end{equation}

\subsubsection{Scalar Encoder}

Each scalar-valued measure \( s^{(k)} \in \mathbb{R} \) is projected into the latent space \( \mathbb{R}^{d} \) through a linear transformation:
\begin{equation}
\mathbf{t}^{(k)} =
\mathbf{W}^{(k)}_s s^{(k)} + \mathbf{b}_s,
\qquad
\mathbf{t}^{(k)} \in \mathbb{R}^{d}.
\end{equation}
where \( \mathbf{W}_s \in \mathbb{R}^{d \times 1} \) and \( \mathbf{b}_s \in \mathbb{R}^{d} \).

This transformation enables consistent dimensionality with high-dimensional inputs during subsequent multimodal interaction.

\subsection{Transformer Encoder for Cross-Modal Interaction}
\label{sec:trans}

We employ a transformer encoder here to capture interactions among heterogeneous measures, enabling the model to learn joint representations that reflect coordinated biological processes underlying BED. Specifically, after the measure tokenization, all measure tokens are concatenated to form a token sequence
\begin{equation}
\mathbf{Z}_0 = [\mathbf{t}^{(1)}, \dots, \mathbf{t}^{(T)}] \in \mathbb{R}^{T \times d},
\end{equation}
where $T = M + K$.

To capture the interdependencies among heterogeneous measures, the token sequence is processed by a transformer encoder which is a stack of $L$ transformer layers,
\begin{equation}
\mathbf{Z}_\ell = \mathrm{TransEnc}_\ell(\mathbf{Z}_{\ell-1}),
\quad \ell = 1,\dots,L,
\end{equation}
with $\mathbf{Z}_L = [\mathbf{z}^{(1)}, \dots, \mathbf{z}^{(T)}] \in \mathbb{R}^{T \times d}$ denoting the representation of output token sequence.
\subsection{Token-wise Mixture-of-Experts}
\label{sec:moe}

Although transformer encoder enables the interaction among all tokens, heterogeneous measures may still exhibit distinct nonlinear response patterns that are not optimally represented through a uniform processing pathway. To address this, we introduce a Mixture-of-Experts module that allows measures tokens to be adaptively processed through specialized transformations in a data-driven manner.

For each modality token representation $\mathbf{z}^{(t)}$ obtained from the transformer encoder, a gating network computes expert assignment weights:
\begin{equation}
\boldsymbol{\alpha}_t
=
\mathrm{softmax}\!\left(
\frac{\mathbf{W}_g \mathbf{z}^{(t)} + \mathbf{b}_g}{\tau_e}
\right),
\end{equation}
where $\tau_e$ is a temperature parameter that controls the sharpness of expert routing.

Each expert is parameterized as a lightweight multilayer perceptron (MLP):
\begin{equation}
\mathrm{Expert}_e(\mathbf{z})
=
\mathbf{W}_2^{(e)}
\sigma\!\left( \mathbf{W}_1^{(e)} \mathbf{z} + \mathbf{b}_1^{(e)} \right)
+ \mathbf{b}_2^{(e)}.
\end{equation}

The expert-refined token representation is then obtained as a weighted aggregation:
\begin{equation}
\mathbf{u}^{(t)}
=
\sum_{e=1}^{E} \alpha_{t,e} \,
\mathrm{Expert}_e(\mathbf{z}^{(t)}).
\end{equation}

This formulation enables the model to automatically group tokens with similar representational characteristics and route them through appropriate expert transformations, thereby improving representation capacity for heterogeneous modalities.

\subsection{Token-Importance Weighting}
\label{sec:tim}

Given the heterogeneous nature of multimodal measurements, different measure may contribute unequally to the final prediction, and their relevance may vary across individuals. To enable subject-specific weighting of measure contributions, we introduce a learnable token-importance module that adaptively assigns importance scores to each token.

Formally, for each refined token representation $\mathbf{u}^{(t)}$, an importance score is computed as
\begin{equation}
\beta_t = \mathbf{w}^\top \mathbf{u}^{(t)} + b,
\end{equation}
which is subsequently normalized via a temperature-controlled softmax function:
\begin{equation}
\pi_t =
\frac{\exp(\beta_t / \tau_p)}
     {\sum_{j=1}^T \exp(\beta_j / \tau_p)}.
\end{equation}

The final pooled representation is then obtained as a weighted aggregation of all measure tokens:
\begin{equation}
\mathbf{u}_{\mathrm{pool}}
=
\sum_{t=1}^{T} \pi_t \, \mathbf{u}^{(t)}.
\end{equation}

This formulation enables adaptive integration of multimodal information while providing interpretable estimates of measure-specific contributions to the prediction.

A two-layer MLP is subsequently employed to map the pooled representation to the classification logits:
\begin{equation}
\boldsymbol{\ell}
=
\mathbf{W}^{(2)}_c
\,
\sigma\!\big(
\mathbf{W}^{(1)}_c \mathbf{z}_{\mathrm{pool}}
+
\mathbf{b}^{(1)}_c
\big)
+
\mathbf{b}^{(2)}_c.
\end{equation}

The entire network is optimized in an end-to-end manner using the cross-entropy loss function.

\section{Experiments}

\subsection{Dataset}

In this study, we adopted the data from the baseline release (Release 4.0) of the ABCD Study~\citep{casey2018adolescent}, a large-scale, multi-site longitudinal initiative compromising over 11,000 children aged 9 to 10 years recruited from 21 sites across the United States. A major advantage of the ABCD Study dataset is its sociodemographic variation, which enhances the generalizability of findings in the U.S. population~\citep{garavan2018recruiting}. Study protocols were approved by the Institutional Review Board (IRB) at the University of California, San Diego, and respective data collection sites, with informed consent and assent obtained from all guardians and participants. As this work is a secondary analysis of fully de-identified data, additional IRB approval was not required.

Following the procedures described in \citep{martin2025distinct}, we constructed a BED dataset comprising 99 individuals with BED and 123 healthy controls. We then extracted multimodal data spanning neuroimaging, behavioral, neuroendocrine, and demographic domains. Specifically, sMRI–derived measures comprise cortical thickness and cortical volume (151 regions), and subcortical volume (14 regions). DTI measures included fractional anisotropy derived from 42 white matter tracts identified using AtlasTrack to quantify white matter tract integrity. Three tfMRI paradigms were used to probe cognitive and reward-related processes: the N-back task, the Monetary Incentive Delay (MID) task, and the Stop Signal Task (SST). Regional topological properties,including nodal efficiency, nodal degree, and nodal betweenness-centrality, were computed across the entire run of trials of the three tasks, yielding 164 nodal measures from the N-back task, 165 from the MID task, and 186 from the SST, as described in~\citep{martin2024distinct,martin2025distinct,martin2025altered}. Behavioral measures were derived from three tfMRI paradigms and include reaction time variability and task accuracy. To capture neuroendocrine influences relevant to development and sex-specific risk, we incorporated hormone measures, including estradiol (ERT), dehydroepiandrosterone (DHEA) and estrone (HSE). Demographic measures include age, sex, household income, BMI percentile, and verbal cognitive ability, as assessed by the Picture Vocabulary Test.


All measures were normalized prior to model training using modality-specific strategies. For high-dimensional neuroimaging measures (e.g., cortical thickness), subject-level z-score normalization was applied across regions or features within each modality, ensuring that values reflected relative spatial patterns rather than absolute magnitude differences. Missing values were imputed with zeros after normalization. For lower-dimensional non-imaging variables, scale adjustments were applied based on their measurement units. For example, reaction time measures were converted from milliseconds to seconds. Hormonal measures were rescaled according to expected value ranges to reduce scale disparities. Demographic measures were also normalized, with verbal IQ scores scaled to a 0-1 range. Any missing scalar values were set to zero following rescaling. After preprocessing and normalization, the dataset was randomly divided at the subject level, with 90\% of the subjects allocated to the training dataset and the remaining 10\% reserved as testing dataset.

\subsection{Implementation Details}

The IMA-MoE model was implemented in \texttt{PyTorch}. Each modality is projected into a shared embedding space with dimensionality $d_{\text{model}} = 128$. The transformer encoder has three transformer layers with one attention head. The MoE module has four experts for dynamic modality-specific routing. The model was trained using the AdamW optimizer with a learning rate of $1 \times 10^{-4}$ and weight decay of $1 \times 10^{-4}$. A linear warm-up over the first five epochs is followed by cosine learning rate decay, for a total of 50 training epochs with a batch size of 16. All experiments were conducted with an NVIDIA RTX 5090 GPU. The model from the final training epoch was used for evaluation on the testing dataset.

\subsection{Prediction Performance of IMA-MoE}
We evaluate the proposed IMA-MoE framework on the testing dataset to assess its ability to differentiate individuals with BED from healthy controls. Model performance is quantified using accuracy, sensitivity, specificity, and F1-score. As summarized in Table~\ref{tab:performance}, the model achieves strong overall performance, with an accuracy of 0.8261 and an F1-score of 0.8571. Notably, the sensitivity (0.9231) is higher than the specificity (0.7000), suggesting that the model is particularly effective at detecting individuals with BED while maintaining reasonable control discrimination.

\begin{table}[ht]
\centering
\caption{Prediction performance of IMA-MoE on testing dataset}
\label{tab:performance}
\begin{tabular}{lcccc}
\hline
\textbf{Group} & \textbf{Accuracy} & \textbf{Sensitivity} & \textbf{Specificity} & \textbf{F1-score} \\
\hline
Overall & 0.8261 & 0.9231 & 0.7000 & 0.8571 \\
Male    & 0.7273 & 0.8571 & 0.5000 & 0.8000 \\
Female  & 0.9167 & 1.0000 & 0.8333 & 0.9231 \\
\hline
\end{tabular}
\end{table}

To further examine potential sex-specific patterns, we evaluate performance separately in male and female subgroups. The model demonstrates higher predictive performance in females than in males across all metrics. In females, the model achieves an accuracy of 0.9167 and a sensitivity of 1.0. Specificity in females also remaines relatively high (0.8333), resulting in an F1-score of 0.9231. In contrast, performance in males is more modest, with an accuracy of 0.7273 and a specificity of 0.5000, suggesting greater difficulty in distinguishing control participants in this subgroup.

\subsection{Modality Contribution Analysis via Token Importance}

To better understand how different measures contribute to model predictions, we analyze the token-importance scores learned by the IMA-MoE framework. Because each measure is represented as a distinct token in our architecture, the token-importance module provides a direct estimate of the relative contribution of each measure to the final classification. With 15 measure tokens initialized uniformly, deviations from the baseline importance (1/15) indicate how the model redistributes importance across modalities during learning.

\begin{figure*}[ht]
\begin{center}
\includegraphics[width=1\linewidth]{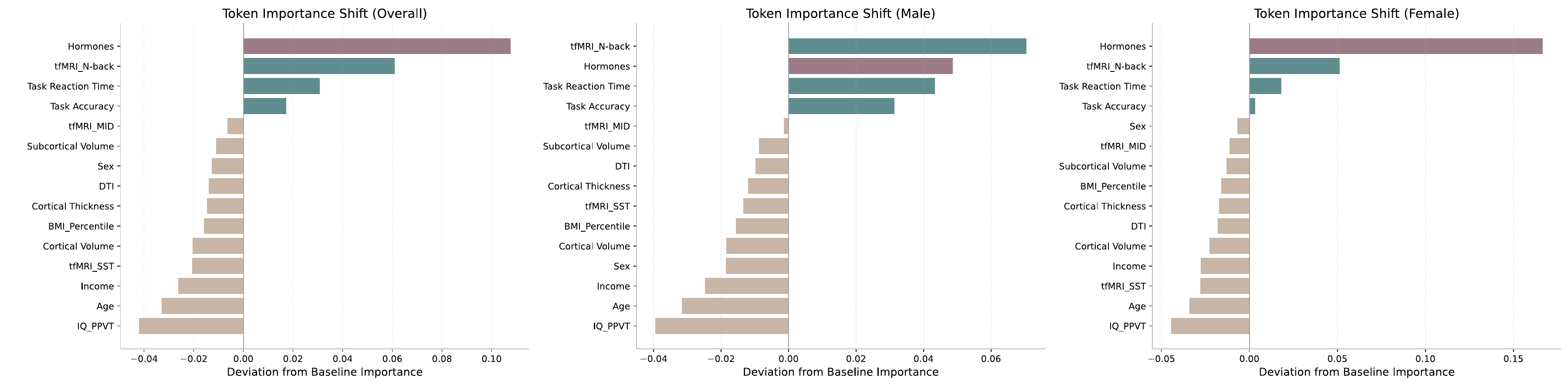}
\end{center}
\caption{Token importance across subjects and sex groups. Group-averaged token-importance values learned by the IMA-MoE model are shown for all subjects (left), males (middle), and females (right). Each bar represents the mean contribution of a modality-specific token to the model’s classification decision. The vertical line indicates the uniform initialization baseline (1/15), and deviations from this baseline reflect how the model redistributes attention across modalities during training.}
\label{fig:fig2}
\end{figure*}

Figure~\ref{fig:fig2} illustrates the average token-importance values across all subjects, as well as separately for males and females. Several clear patterns emerge. The model consistently assigns greater importance to tfMRI tokens and hormones token compared with many structural or demographic tokens. In particular, tokens corresponding to hormonal measures, task performance metrics (e.g., accuracy and reaction time), and N-back tfMRI features show elevated importance relative to the uniform baseline, suggesting that functional and hormonal features play a central role in distinguishing BED from healthy controls. In contrast, several demographic measures and some structural measures remain near or below the baseline, indicating a comparatively smaller influence on the model’s predictions.

Sex-stratified analyses further reveal systematic differences in modality contribution. Although both groups demonstrate strong reliance on tfMRI measures, females exhibit substantially higher importance for hormonal measures, whereas males show relatively greater contributions from N-back tfMRI measures and task performance measures. These findings indicate that the model captures sex-specific patterns in how neurobiological and neuroendocrine factors relate to BED classification, rather than relying on a uniform predictive strategy.

\begin{figure}[t]
\begin{center}
\includegraphics[width=1\linewidth]{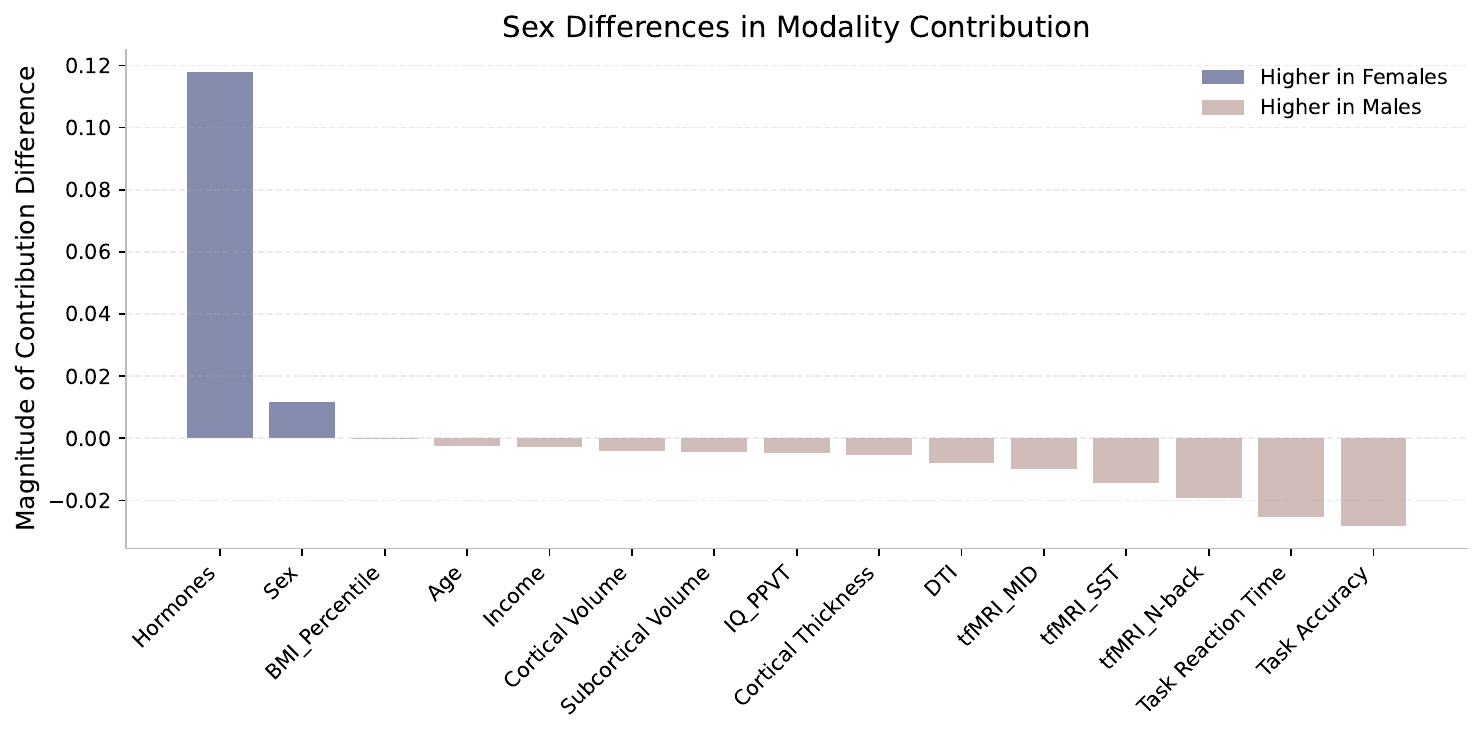}
\end{center}
\caption{Sex differences in measures contribution to model predictions. Bars are ordered based on the signed difference (female minus male), with bar height representing the magnitude of the difference.}
\label{fig:fig3}
\end{figure}

To further highlight these differences, Figure~\ref{fig:fig3} visualizes the sex differences in token importance for each individual token (female minus male). Hormones token display the most pronounced female-skewed contribution, indicating that hormonal measures play a substantially larger role in the model’s predictions for females. Sex itself also shows a modest female-weighted contribution, while BMI percentile exhibits minimal sex difference, suggesting a relatively balanced influence across groups. In contrast, task performance measures and tfMRI features from the N-back, SST, and MID paradigms show higher contributions in males. Structural neuroimaging measures, DTI measures, and demographic measures demonstrate comparatively small differences, indicating that these modalities contribute in a more similar manner across sexes. Overall, the pattern suggests that hormonal measures are particularly informative for females, whereas tfMRI and task performance measures are relatively more informative for males, with many structural and demographic measures showing limited sex-specific divergence in their contributions.

\subsection{Effectiveness of Multimodal Representation Learning}

In this section, we investigate the effectiveness of multimodal representation learning paradigm proposed in this study. We first examine the predictive performance of single modality by training IMA-MoE using structural, functional, behavioral, hormonal, and demographic measures independently. As summarized in Table~\ref{tab:modality_ablation}, models trained on structural or demographic measures alone exhibit high sensitivity but near-zero specificity, indicating a tendency to over-predict BED status in the absence of complementary information. In contrast, model trained on functional measures achieves the strongest unimodal performance (accuracy = 0.7391). Hormonal measures alone also demonstrate moderate discriminative capacity (accuracy = 0.6522). Behavioral measures provide additional but limited predictive value. Collectively, these results indicate that no single modality is sufficient for good predictive performance. Multimodal integration is necessary to capture complementary neurobiological features relevant to disorder expression.

\begin{table}[h]
\centering
\caption{Single-modality prediction results. STR: structural MRI and DTI measures; 
FUN: tfMRI measure; HORM: hormonal measure; 
BEH: behavioral task performance measures; DEMO: demographic measures.}
\label{tab:modality_ablation}
\begin{tabular}{lcccc}
\hline
Modality & Acc & Sens & Spec & F1 \\
\hline
STR  & 0.5652 & 1.0000 & 0.0000 & 0.7222 \\
HORM & 0.6522 & 0.8462 & 0.4000 & 0.7333 \\
DEMO & 0.5652 & 1.0000 & 0.0000 & 0.7222 \\
FUN  & 0.7391 & 0.9231 & 0.5000 & 0.8000 \\
BEH  & 0.6087 & 0.9231 & 0.2000 & 0.7273 \\
\hline
\end{tabular}
\end{table}

To further evaluate the effectiveness of mulitmodal data fusion design in IMA-MoE, we compare it with several machine leanring and deep learning baselines trained on the same flattened multimodal inputs. Specifically, we implement linear Support Vector Machine (SVM), radial basis function (RBF) SVM, Random Forest, and a five-layer multilayer perceptron (MLP), each of which received the identical concatenated feature vector comprising structural, functional, hormonal, behavioral, and demographic measures.

\begin{table}[h]
\centering
\caption{Comparison with machine learning and deep learning baselines trained on flattened multimodal measures.}
\label{tab:baseline_comparison}
\begin{tabular}{lcccc}
\hline
Model & Acc & Sens & Spec & F1 \\
\hline
Linear SVM     & 0.6957 & 0.6923 & 0.7000 & 0.7200 \\
RBF SVM        & 0.6957 & 1.0000 & 0.3000 & 0.7879 \\
Random Forest  & 0.6522 & 1.0000 & 0.2000 & 0.7647 \\
MLP (5-layer)  & 0.6957 & 0.8462 & 0.5000 & 0.7586 \\
IMA-MoE        & \textbf{0.8261} & \textbf{0.9231} & \textbf{0.7000} & \textbf{0.8571} \\
\hline
\end{tabular}
\end{table}

As summarized in Table~\ref{tab:baseline_comparison}, machine learning models achieve comparable sensitivity but consistently lower specificity compared with the proposed architecture. In particular, RBF SVM and Random Forest exhibite perfect sensitivity but poor specificity (0.3000 and 0.2000, respectively), indicating a tendency to over-predict BED status when relying solely on flattened multimodal features. The five-layer MLP baseline improves specificity relative to classical models, yet still underperforms the IMA-MoE framework across all evaluation metrics. In contrast, IMA-MoE achieves the highest overall performance while maintaining a balanced trade-off between sensitivity and specificity. These results suggest that the performance gains of the proposed framework are not solely attributable to more modalities and increased nonlinear modeling capacity, but rather to its ability to explicitly model modality interactions and measure-specific contributions.

\subsection{Ablation Study}

\begin{table*}[ht]
\centering
\caption{Ablation study examining the contribution of architectural components to classification performance. \checkmark denotes inclusion of the corresponding component. Trans: Transformer encoder; MoE: Mixture-of-Experts module; TIM: token-importance module; Avg: token-wise average pooling for classification.}
\label{tab:model_ablation}
\begin{tabular}{lcccccccc}
\hline
Model & Tokens & Transformer & MoE & Token Imp. & Acc & Sens & Spec & F1 \\
\hline
Token + Avg & \checkmark & $\times$ & $\times$ & $\times$ & 0.7391 & 0.8462 & 0.6000 & 0.7857 \\
Token + MoE + TIM & \checkmark & $\times$ & \checkmark & \checkmark & 0.6522 & 0.6923 & 0.6000 & 0.6923 \\
Token + Trans + TIM & \checkmark & \checkmark & $\times$ & \checkmark & 0.7826 & 0.9231 & 0.6000 & 0.8276 \\
Token + Trans + Avg & \checkmark & \checkmark & \checkmark & $\times$ & 0.8261 & 0.9231 & 0.7000 & 0.8571 \\
IMA-MoE & \checkmark & \checkmark & \checkmark & \checkmark & \textbf{0.8261} & \textbf{0.9231} & \textbf{0.7000} & \textbf{0.8571} \\
\hline
\end{tabular}
\end{table*}

To assess the contribution of key architectural components to classification performance, we conducted a series of ablation experiments by systematically removing the transformer encoder, MoE module, and token-importance module from the IMA-MoE framework. As shown in Table~\ref{tab:model_ablation}, removal of the Transformer encoder markedly degrades performance across all metrics, suggesting that modeling dependencies among heterogeneous modality representations is crucial to discriminate BED and healthy controls.
Removal of the MoE module also yielded the downgraded performance, decreasing accuracy from 0.8261 to 0.7826 and specificity from 0.7000 to 0.6000. Notably, the inclusion of the token-importance module did not degrade classification performance, suggesting that interpretability can be enhanced without sacrificing predictive capacity. Overall, these findings indicate that all components of the IMA-MoE framework are essential for accurate discrimination between BED and healthy controls.

\section{Discussion}

The IMA-MoE framework demonstrated robust overall classification performance (Accuracy = 82.61\%; F1 = 0.8571), exceeding the random chance baseline of 50\% by a large margin. This result indicates that the model effectively synthesizes heterogeneous multimodal measures to identify the latent patterns that distinguish BED from healthy controls. However, the sex-stratified analysis revealed a stark disparity, with the model achieving better predictive performance in females (Accuracy = 91.67\%) compared to males (Accuracy = 72.73\%). This suggests that the features most discriminative for females may be less informative for males, likely rooted in sexual differences in clinical presentation and underlying neurobiology~\citep{reslan2011college,murray2017enigma,murray2023sex}. For example, males and females exhibit differences in their conceptualization of binge episodes, with males more likely to emphasize the amount of food consumed and females more likely to report a loss of control during eating~\citep{reslan2011college}. Recent neuroimaging evidence indicates that girls with BED exhibit unique elevations in gray matter density in regions related to reward and cognitive control~\citep{murray2023sex1}. Furthermore, functional connectivity analyses have identified distinct sex-specific alterations, such as boys with BED displaying wider-spread hypoconnectivity in posterior reward regions compared to the more localized prefrontal deficits seen in girls~\citep{murray2023sex}. Thus, the current multimodal measures may not fully capture male-specific alterations associated with BED. More discriminative measures remain to be explored to better characterize the underlying divergent neurobiological mechanisms across sexes.

Our token-importance analysis further reveals a fundamental dissociation in the most predictive features of BED, suggesting distinct etiological pathways for males and females. IMA-MoE's heavy weighting of the hormonal tokens suggests that for females, the disorder is inextricably linked to the dynamic neuroendocrine background, indicating that hormonal measures may serve as potential biomarkers for elucidating female-specific neurobiological mechanisms underlying BED. It also offers computational support for the "activational" hormone hypothesis, which posits that pubertal fluctuations in ovarian hormones modulate neural and genetic risk for binge eating~\citep{klump2017sex,klump2018estrogen,klump2020disruptive}. This aligns with extensive literature indicating that lower levels of estradiol, particularly during the mid-luteal or premenstrual phases, can "turn on" genetic risks for dysregulated eating and amplify reward sensitivity in vulnerable females~\citep{klump2006preliminary,klump2017sex,mikhail2021gonadal}. Such mechanism is largely absent in males due to the protective organizational effects of perinatal testosterone and different pubertal hormonal milieus~\citep{klump2020disruptive,mikhail2021gonadal}. In contrast, the male-specific reliance on tfMRI measures (N-back, SST, MID) suggests that BED pathology in males may be more associated with deficits in inhibitory control and reward processing, relatively independent of cyclic neuroendocrine modulation seen in females. This interpretation is consistent with reports that males with BED exhibit distinct neurofunctional profiles, including alterations in attentional and inhibitory control networks that are less directly influenced by hormonal fluctuations~\citep{murray2023sex,martin2025altered}.

Interestingly, structural MRI and demographic tokens contributed minimally to the model’s decision-making process, remaining near or below baseline importance. This finding adds to the ongoing debate regarding structural brain alterations in BED, for which the literature remains inconsistent: while some studies report gray matter volume differences in reward-related regions such as the medial orbitofrontal cortex~\citep{schafer2010regional} and left nucleus accumbens~\citep{abdo2020relationship}, others have failed to replicate these findings~\citep{abdo2020relationship,hagan2021subcortical}. The low importance assigned to structural tokens in our IMA-MoE model suggests that BED may be more appropriately conceptualized as a functional connectopathy, characterized by aberrant communication between reward and inhibitory control networks, rather than as a disorder defined by stable structural abnormalities. Furthermore, the limited contribution of demographic variables, particularly BMI, indicates that the IMA-MoE model captured neurobiological signatures of psychopathology beyond anthropometric correlates of obesity. 

\section{Conclusion}
In this study, we proposed an IMA-MoE framework for integrating heterogeneous neuroimaging, behavioral, endocrine, and demographic measures associated with BED. By preserving modality-specific information while enabling cross-modal interaction and adaptive expert specialization, the proposed architecture achieves superior classification performance compared with baseline approaches. Importantly, the token-importance mechanism provides interpretable estimates of measure-level contribution, facilitating data-driven insights into the factors contributing to BED classification. Overall, these findings highlight the promise of modality-aware, interpretable multimodal learning for advancing the characterization of BED neurobiology.

\section*{Acknowledgements}

This work was partially supported by NIMH (R01 MH126448) and New Jersey State Department of Health (CBIR25IRG001). Data used in  preparation of this article were obtained from the Adolescent Brain Cognitive Development (ABCD) Study (https://abcdstudy.org), held in the NIMH Data Archive (NDA). This is a multisite, longitudinal study designed to recruit more than 10,000 children aged 9-10 and follow them over 10 years into early adulthood. The ABCD Study® is supported by the National Institutes of Health and additional federal partners under award numbers U01DA041048, U01DA050989, U01DA051016, U01DA041022, U01DA051018, U01DA051037, U01DA050987, U01DA041174, U01DA041106, U01DA041117, U01DA041028, U01DA041134, U01DA050988, U01DA051039, U01DA041156, U01DA041025, U01DA041120, U01DA051038, U01DA041148, U01DA041093, U01DA041089, U24DA041123, U24DA041147. A full list of supporters is available at https://abcdstudy.org/federal-partners.html. A listing of participating sites and a complete listing of the study investigators can be found at \url{https://abcdstudy.org/consortium_members/}. ABCD consortium investigators designed and implemented the study and/or provided data but did not necessarily participate in the analysis or writing of this report. This manuscript reflects the views of the authors and may not reflect the opinions or views of the NIH or ABCD consortium investigators. The ABCD data repository grows and changes over time. The ABCD data used in this report came from 10.15154/h3rr-zp41.

\bibliography{ref}
\bibliographystyle{elsarticle-harv.bst}

\end{document}